\pdfoutput=1

\documentclass[11pt]{article}

\usepackage{EMNLP2022}

\usepackage{times}
\usepackage{latexsym}
\usepackage{algorithm}
\usepackage{algpseudocode}
\usepackage[T1]{fontenc}

\usepackage[utf8]{inputenc}

\usepackage{microtype}

\usepackage{inconsolata}

\usepackage{comment}
\usepackage{bm}

\usepackage{multirow}
\usepackage{makecell}
\usepackage{textcomp}
\usepackage{graphicx}

\usepackage{makecell}
\usepackage{longtable}
\usepackage{subcaption}
\usepackage{lipsum}
\usepackage{multicol}
\title{SMSMix: Sense-Maintained Sentence Mixup for Word Sense Disambiguation}

\author{Hee Suk Yoon$^{1}$\Thanks{ Equal contribution}, Eunseop Yoon$^{1}$\footnotemark[1], John Harvill$^{2}$, Sunjae Yoon$^{1}$, \\ 
\bf{Mark Hasegawa-Johnson}$^{2}$, \bf{and Chang D. Yoo}$^{1}$\Thanks{ Corresponding author} \\
         $^{1}$Korea Advanced Institute of Science and Technology (KAIST) \\
         $^{2}$University of Illinois at Urbana-Champaign (UIUC) \\ \texttt{\{hskyoon, esyoon97, sunjae.yoon, cd\_yoo\}@kaist.ac.kr} \\ \texttt{\{harvill2, jhasegaw\}@illinois.edu}}

\begin{document}
\maketitle
\begin{abstract}
Word Sense Disambiguation (WSD) is an NLP task aimed at determining the correct sense of a word in a sentence from discrete sense choices. Although current systems have attained unprecedented performances for such tasks, the nonuniform distribution of word senses during training generally results in systems performing poorly on rare senses. To this end, we consider data augmentation to increase the frequency of these least frequent senses (LFS) to reduce the distributional bias of senses during training. We propose Sense-Maintained Sentence Mixup (SMSMix), a novel word-level mixup method that maintains the sense of a target word. SMSMix smoothly blends two sentences using mask prediction while preserving the relevant span determined by saliency scores to maintain a specific word's sense. To the best of our knowledge, this is the first attempt to apply mixup in NLP while preserving the meaning of a specific word. With extensive experiments, we validate that our augmentation method can effectively give more information about rare senses during training with maintained target sense label.
\end{abstract}

\section{Introduction}

Determining the meaning of a word in a particular sentence is a fundamental problem in Natural Language Processing (NLP), as it can enable a better understanding of natural languages and help solve various NLP problems. Word Sense Disambiguation (WSD) is a crucial task towards accomplishing this goal, where it involves choosing the most relevant meaning of a target word in context from predefined sense labels. Like most other NLP tasks, the advancement of Deep Learning has led supervised learning of neural models to be the primary method of WSD \cite{GlossBERT, BEM, ESC}. However, one of the biggest challenges of WSD is overcoming the data bias that naturally stems from the distributional bias of senses in language \cite{zipfian}.
\begin{figure}[t]
	\centering
	\includegraphics[width=\linewidth]{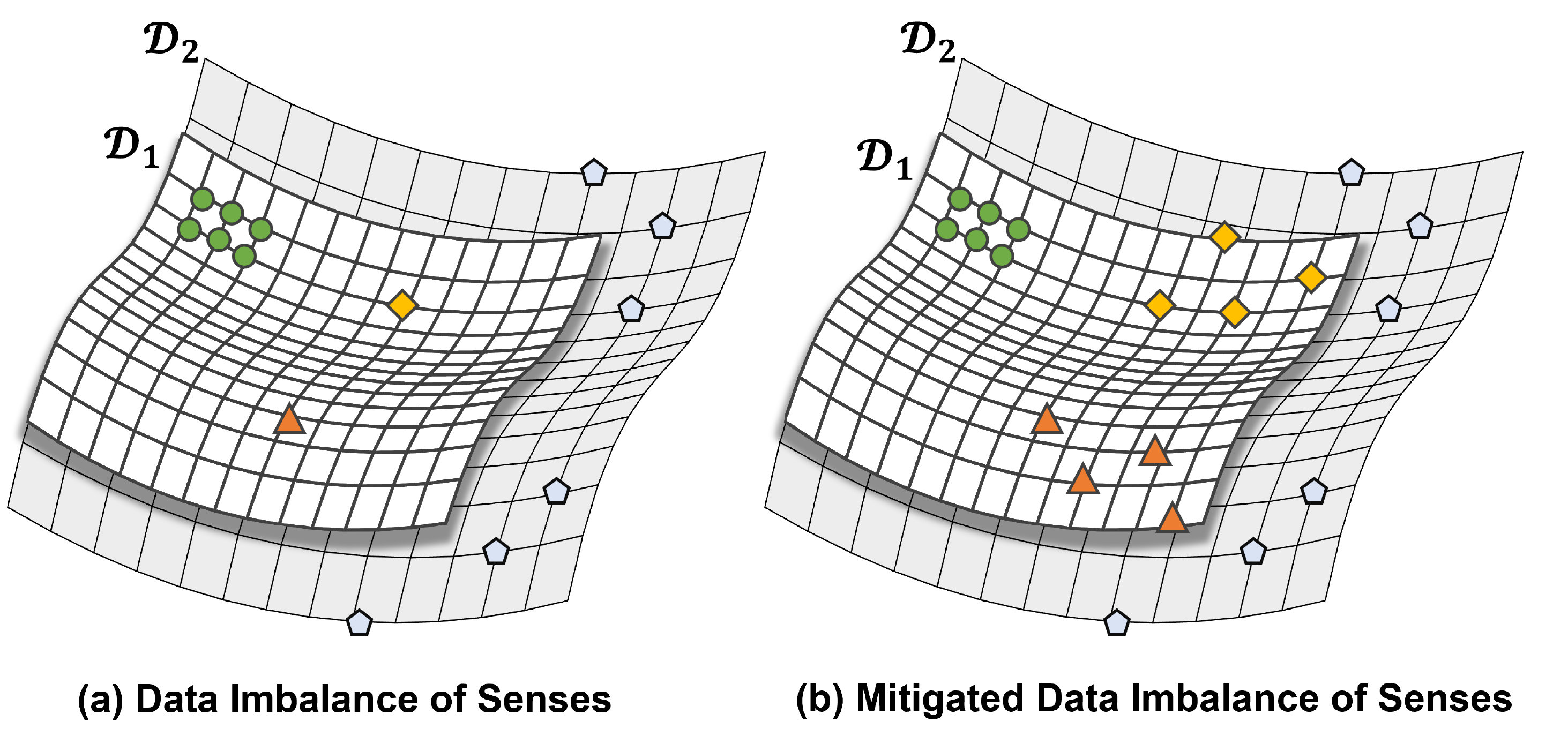}
	\caption{Schematic illustration of SMSMix filling underpopulated areas in the training data space ($\mathcal{D}_1$) - linked to least frequent senses (LFS) - with synthetic examples created through mixup with external corpus ($\mathcal{D}_2$) sentences.}
	\label{fig:1}
\end{figure}
\begin{figure*}[t]
	\centering
	\includegraphics[width=\linewidth]{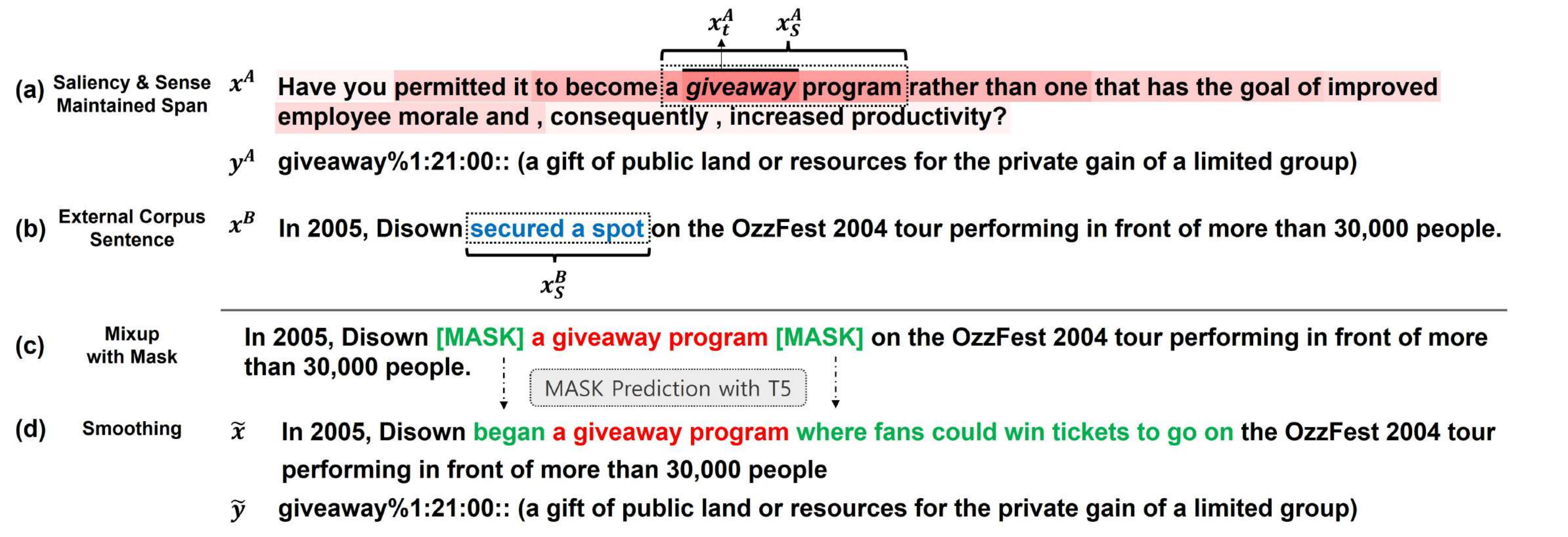}
	\caption{Overview of SMSMix. 
	(a) $x^A$ is a sentence containing target word $x_t^A$ with sense label $y^A$. The saliency score of each token is visualized by the concentration of the color.
	The sense-maintained span $x_S^A$ is determined by setting the highest saliency token from the front and back of the target word $x_t^A$ as the start and end of the span, respectively. (b) A random sentence $x^B$ with random span $x_S^B$ is sampled from an external corpus. (c) $x_S^A$ is replaced into $x_S^B$ with mask tokens. (d) Mask token prediction with T5 smoothly blends the two sentences to minimize the strong-edge problem. The output results in $\tilde{x}$ containing the target word $x_t^A$ with label $\tilde{y}$ set to $y^A$.}
	\label{fig:2}
\end{figure*}
Because the dataset tends to have this bias toward certain senses, the WSD system shows high accuracy on the most frequent sense (MFS) and low accuracy on the least frequent sense (LFS) of a word. 

A common solution in machine learning to address data imbalance involves oversampling under-represented categories, although merely doing so often leads to overfitting of minority classes \cite{oversampling}. Another approach uses data augmentation to drive the learning process toward a more suitable solution. Due to the expensive cost of data collection, data augmentation has become one of the essential tools in modern deep learning, especially when dealing with low-resource tasks. In image classification, some approaches \cite{Remix, MixBoost, BalancedMixup} have been recently explored, involving the well-known data augmentation method Mixup \cite{mixup} to alleviate data imbalance. This motivation encourages us to fill underpopulated areas in the training data of WSD - linked to least frequent senses (LFS) - with synthetic examples created through mixup (Figure \ref{fig:1}).



In this work, we propose Sense-Maintained Sentence Mixup (SMSMix), a novel mixup data augmentation method where the meaning of a specific word in a sentence is unaffected as much as possible while exposing it to various contexts. 
Inspired by the recently proposed word-level mixup in NLP for text classification tasks \cite{SSMix}, we first determine a span of text in a sentence containing a target word that is most relevant to maintaining the target word sense using gradient-based saliency scores. Second, because \citet{LinearTransformationDataAug} mentions that mixup within training data cannot give more information to the model and acts merely as a regularizer, we inject the aforementioned span of text into a random sentence from a Wikipedia corpus. Lastly, when carrying out this injection, we consider smoothly blending the span to the Wikipedia corpus sentence using mask prediction, inspired by SmoothMix \cite{SmoothMix}, where they consider smoothly mixing two images to solve the strong-edge problem.

SMSMix has empirically shown to be an effective augmentation method that can give more information about rare senses during training a WSD model in a supervised setting. We especially show that performing mixup with an external sentence (i.e., a sentence that is not from the training data) can outperform internal mixup within training data. Also, by visualizing the latent vector of the target words in the augmented sentences, we show that SMSMix effectively maintains the sense of the target word.


\section{SMSMix}

SMSMix generates a new sentence $\tilde{x}$ by injecting span $x_S^A$, which contains a target word $x_t^A$ with sense label $y^A$, from sentence $x^A$ into another span $x_S^B$ from sentence $x^B$. Prior to injection, we concatenate mask tokens to the front and back of span $x_S^A$ and perform mask prediction to smoothly blend the two sentences. 
Because we use saliency score to preserve the sense label when determining span $x_S^A$, we set the label of $x_t^A$ in the new sentence $\tilde{x}$ to $y^A$  (Figure~\ref{fig:2}).
\subsection{Saliency and Sense-Maintained Span}
Saliency, which shows how each input fraction affects the final prediction, is usually measured using gradient-based methods.
\citet{SSMix} recently proposed a mixup strategy in NLP using a gradient-based saliency score to preserve the locality of the two sentences performing mixup. 
Similarly, we compute the gradient of classification loss $L$ with respect to the input token embedding $e$, and apply the L2 norm to get the saliency for each input token: i.e., $s=||\frac{\partial L}{\partial e}||_2$. 
The saliency of each input token signifies how influential each input token is for the meaning of the target word. The most salient span for maintaining the target word sense is determined by taking the token with the highest saliency score at the front and back of the target word and setting it to the beginning and the end of the span, respectively.  

\subsection{Mixing Sentence}
We consider two scenarios when injecting the sense-maintained span. First, we consider injecting the span into a random MFS sentence from the training set for the corresponding sense (internal). In this case, $x_S^B$ is determined in the same way as $x_S^A$. However, \citet{LinearTransformationDataAug} mentions that mixup inside the training data has limitations in giving new information to the model and is effective only as a regularization role. Thus, we also consider injecting the span into a random sentence from the Wikipedia corpus (external). 

\subsection{Smoothing}
SmoothMix \cite{SmoothMix} has been shown to be an effective mixup strategy in image classification for minimizing the problem of unnatural strong-edge of the boundary between two images performing the mixup. Motivated by SmoothMix, we propose to minimize the strong-edge problem in the boundary between the two contributing sentences. Prior to injecting the sense-maintained span onto the other sentence, we put a mask token at the front and back of the span. We then use T5 \footnote{We use T5 Version 1.1-large for all the mask predictions. \href{https://huggingface.co/google/t5-v1\_1-large}{https://huggingface.co/google/t5-v1\_1-large}} \cite{T5} for mask prediction to generate a smoothly transitioning mask and use this to blend two different texts to form an augmented sample.

\section{Experimental Setup}
\subsection{Dataset and Model}
\paragraph{Dataset} We use the SemCor \cite{semcor} for training, the largest dataset manually annotated with sense from WordNet that contains 226,036 annotated examples covering 33,362 unique senses. As standard procedure, we use SemEval-2007 (\textbf{SE07}, \citet{se07}) as the validation set, while performing testing on Senseval-2 (\textbf{SE2}; \citet{SE2}), Senseval-3 (\textbf{SE3},  \citet{SE3}), SemEval-2013 (\textbf{SE13}, \citet{SE13}), and SemEval-2015 (\textbf{SE15}, \citet{SE15}). Additionally, to measure the degree to which the system generalizes to LFS, unseen words, and definitions (zero-shot settings), we consider five subsets of the data (MFS, LFS, 0-lex, 0-lex-def, 0-def) proposed in \citet{ESC}.

\paragraph{Model} We evaluate our augmentation scheme on the BEM \cite{BEM} system, which employs a bi-encoder to represent the target word and its sense definitions within the same space. 
\subsection{Training}
We adopt a two-stage training strategy as \citet{SSMix} and \citet{twostep}. After fully training the WSD system, we train one additional epoch with data augmentation with a learning rate of 5e-7. All other hyperparameters are set equal to the original BEM training \footnote{\href{https://github.com/facebookresearch/wsd-biencoders}{https://github.com/facebookresearch/wsd-biencoders}}. 
We consider three types of data augmentation: (1) simple oversampling of LFS (Oversample), (2) SMSMix with internal training data sentence (SMSMix\textsubscript{\textit{int}}), (3) SMSMix with random external Wikipedia corpus sentence (SMSMix\textsubscript{\textit{ext}}). Additionally, for both settings of SMSMix, we automatically filter out those augmented sentences that are not grammatically acceptable, which could happen when the T5 could
not smoothly blend the span into a sentence \footnote{We use the sentence acceptability judgment capability of T5-large using the "cola sentence:" prompt.}. For all three data augmentations, we add three data samples for half of the LFS in the SemCor training data. 
When training with data augmentation, we concatenate the original data with the augmented data to prevent the training data distribution from getting too far from the original data distribution, as mentioned in \citet{distributionGap}. All training was done on NVIDIA Quadro RTX 8000.
\subsection{Evaluation}
Performance of the WSD task has generally been reported so far in terms of micro-average F1 scores. However, doing so gives more weight to frequent senses simply because they occur more often \footnote{Out of 6,368 test data in the aggregated ALL evaluation set, 4949 are MFS and 1419 are LFS.}, thus resulting in an underrepresentation of the low performances of the least frequent senses. Therefore, in addition to the micro-averaged F1 scores, we also choose to report the macro-averaged F1 scores as \citet{microF1}. 

\begin{table*}[t]
	\centering
	\begin{tabular}{l||c c| c c c c c c c c c c}
		\Xhline{3\arrayrulewidth}
		& \multicolumn{2}{c|}{Dev Set} & \multicolumn{10}{c}{Test Sets} \\
		\cline{2-13}
		Methods                              & \multicolumn{2}{c|}{SE07}     & \multicolumn{2}{c}{SE2}     &\multicolumn{2}{c}{SE3}     & \multicolumn{2}{c}{SE13}     & \multicolumn{2}{c}{SE15}   & \multicolumn{2}{c}{ALL}     \\  
		                                    &\begin{small}m-F1\end{small} &\begin{small}M-F1\end{small}&\begin{small}m-F1\end{small} &\begin{small}M-F1\end{small}&\begin{small}m-F1\end{small} &\begin{small}M-F1\end{small}&\begin{small}m-F1\end{small} &\begin{small}M-F1\end{small}&\begin{small}m-F1\end{small} &\begin{small}M-F1\end{small}&\begin{small}m-F1\end{small} &\begin{small}M-F1\end{small}\\ 
		\Xhline{2\arrayrulewidth}
		\begin{small}BEM\end{small}     & \bf{74.5} & \bf{72.4} & 79.4 & 75.5 & 77.4 & 73.0 & 79.7 & 77.0 & 81.7 & 79.1 & 79.0 & 73.9    \\
		\begin{small} +Oversample\end{small}   & 74.1 & 72.0 & 79.6 & 75.8 & 77.7 & 73.5 & 79.7 & \bf{77.7} & 81.8 & 79.4 & 79.1 & 74.6 \\ 
		\begin{small} +SMSMix$_{int}$\end{small} & 74.3 & 72.0 & 79.4  & 75.7 & \bf{77.9} & \bf{73.8} & 79.3 & 77.5 & 81.9 & 79.1 & 79.1 & 74.4\\
		\begin{small} +SMSMix$_{ext}$\end{small} & \bf{74.5} & \bf{72.4} & \bf{79.9} & \bf{76.4} & 77.8 & \bf{73.8} & \bf{79.8} & \bf{77.7} & \bf{82.2} & \bf{79.6} & \bf{79.3} & \bf{74.8}
		\\\hline

		\Xhline{3\arrayrulewidth} 

	\end{tabular}
	\caption{Comparison of both micro (m-F1) and macro (M-F1) F1 scores on the all-words WSD task against its baseline (BEM). The reported values are an average over five runs with different seeds (The standard deviations are reported in Appendix \ref{tab:overallstd}). Best macro and micro F1 scores for each test set are in \textbf{bold}.}
	\label{tab:overall}
\end{table*}
\begin{table*}[t]
	\centering

	\begin{tabular}{l|| c c c c c c c c c c}
		\Xhline{3\arrayrulewidth}
		  &\multicolumn{2}{c}{MFS}     & \multicolumn{2}{c}{LFS}     & \multicolumn{2}{c}{0-lex}   & \multicolumn{2}{c}{0-lex-def}  &\multicolumn{2}{c}{0-def}   \\  
		                                    &\begin{small}m-F1\end{small} &\begin{small}M-F1\end{small}&\begin{small}m-F1\end{small} &\begin{small}M-F1\end{small}&\begin{small}m-F1\end{small} &\begin{small}M-F1\end{small}&\begin{small}m-F1\end{small} &\begin{small}M-F1\end{small}&\begin{small}m-F1\end{small} &\begin{small}M-F1\end{small}\\ 
		\Xhline{2\arrayrulewidth}
		BEM &  \bf{89.3} & 79.0 & 53.3 & 57.7 & 90.2 & 90.3 & 68.2 & 65.7 & 69.1 & 67.1 \\
		+ Oversample & 89.0 & 79.3 & 54.7 & 59.0 & 90.7 & 90.9 & 68.8 & 66.7 & \bf{69.7} & 68.0\\
		+ SMSMix$_{int}$ & 89.2 & 79.2 & 54.6 & 59.3 & 90.5 & 90.9 & 68 & 66.3 & 68.2 & 67.2\\
		+ SMSMix$_{ext}$  & \bf{89.3} & \bf{79.7} & \bf{55.0} & \bf{59.7} & \bf{91} & \bf{91.6} & \bf{69.0} & \bf{66.9} & \bf{69.7} & \bf{68.2}\\

		\Xhline{3\arrayrulewidth}
	\end{tabular}
	\caption{Comparison of both micro (m-F1) and macro (M-F1) F1 scores on MFS, LFS, and zero-shot datasets. The reported values are an average over five runs with different seeds (The standard deviations are reported in Appendix \ref{tab:freqstd}). Best macro and micro F1 scores for each dataset are in \textbf{bold}.}
	\label{tab:freq}
\end{table*}

\section{Results and Discussion}

\subsection{Overall Results}
Table \ref{tab:overall} shows the overall micro (m-F1) and macro F1 (M-F1) results on the English all-words WSD task. We find that oversampling and SMSMix within training data (SMSMix$_{int}$) have a similar slight increase over the original BEM. Performing SMSMix with external Wikipedia corpus (SMSMix$_{ext}$) shows the highest increase in performance, obtaining 79.3 m-F1 and 74.8 M-F1 on the aggregated ALL evaluation set, outperforming the original BEM by 0.3 m-F1 and 0.9 M-F1 points, respectively. 
\subsection{Results on Sense Frequency}
In Table~\ref{tab:freq} we report the results on the five subsets of the data (MFS, LFS, 0-lex, 0-lex-def, 0-def). 
Compared with BEM without data augmentation, the performance on LFS of oversampling, SMSMix\textsubscript{\textit{int}}, and SMSMix\textsubscript{\textit{ext}} improved by 1.3, 1.6, and 2.0 M-F1, respectively. 
As the performance is improved in all three cases, it can be seen that the approach to mitigate sense imbalance through data augmentation is effective. In addition, the results have shown that performing mixup augmentation through external information obtained more performance gain than the internal mixup.
 \begin{figure}[t]
	\centering
	\includegraphics[width=\linewidth]{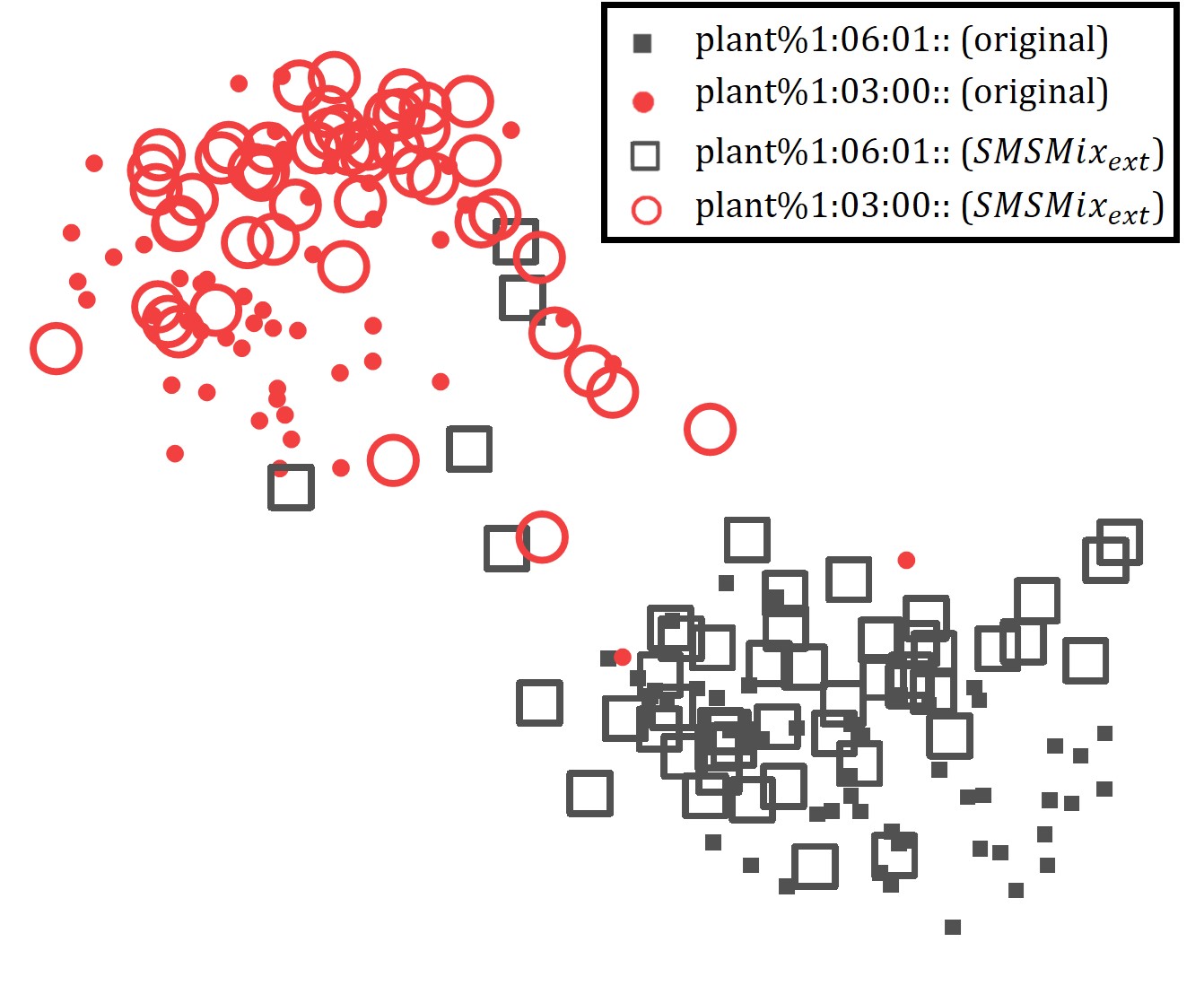}
	\caption{Latent space visualization of the embeddings of target word \emph{plant (noun)} in the labeled OMSTI dataset sentences (original) and augmented sentences using SMSMix on SemCor training dataset. The embeddings of SMSMix target words closely overlap with those of original target words, suggesting that SMSMix maintains the true sense label of the target word.}
	\label{fig:3}
\end{figure}
\subsection{Sense-Maintained Augmentation}
\label{sec:tsne}
 We visually analyze whether SMSMix changes the meaning of the target word. We first take the context encoder of a BEM system fully trained on the SemCor dataset without augmentation. Then, we apply SMSMix to the SemCor training set to generate 50 augmented sentences per sense. Moreover, we obtain 50 labeled sentences per sense from the OMSTI dataset \cite{OMSTI}. These are then fed into the pre-trained BEM system, and we extract the embedding of the target words. We plot the 2-D representation of these embeddings using t-SNE \cite{tsne}. We find that the resulting latent space visualization of the target words in augmented sentences closely overlaps with those in labeled sentences (Figure \ref{fig:3}), which suggests that SMSMix effectively preserves the meaning of the target word while mixing it into various contexts.

\section{Conclusion}
This paper introduced a novel input-level mixup data augmentation scheme SMSMix for improving the Least Frequent Sense (LFS) data imbalance in the Word Sense Disambiguation task.
SMSMix maintains the meaning of a specific word in a sentence by keeping the sense-maintained span using saliency score and smoothly injects the span into a different context using mask prediction. Throughout the experiment, we show that instead of injecting the sense-maintained span with an internal training data sentence, injecting it into a random external corpus sentence allows the model to better improve the performance on LFS. 

\section*{Limitations}
In this paper, we considered word-level mixup data augmentation to create new synthetic sentences containing a specific word with preserved meaning. We show in Appendix \ref{aug_example} that when the two sentences performing the mixup have a high contextual difference, the T5 model fails to smoothly blend the two sentences during mask prediction, resulting in a sentence that is grammatically incorrect or does not make sense. For future work, we plan on considering sentence similarity to choose sentences for mixup instead of random selection, as in the paper. 
\section*{Ethics Statement}
This paper does not violate the use of others' work without reference. Furthermore, the paper does not involve introducing new datasets and the experiments conducted do not utilize demographic or identity characteristics.

\section*{Acknowledgements}
This work was supported by Institute of Information \& communications Technology Planning \& Evaluation (IITP) grant funded by the Korea government (MSIT) (No. 2022-0-00951, Development of Uncertainty-Aware Agents Learning by Asking Questions), and was partly supported by the National Research Foundation of Korea (NRF) grant funded by the Korea government(MSIT) (No. 2022R1A2C201270611).

\bibliography{main}

\begin{thebibliography}{24}
\expandafter\ifx\csname natexlab\endcsname\relax\def\natexlab#1{#1}\fi

\bibitem[{Barba et~al.(2021)Barba, Pasini, and Navigli}]{ESC}
Edoardo Barba, Tommaso Pasini, and Roberto Navigli. 2021.
\newblock \href {https://doi.org/10.18653/v1/2021.naacl-main.371} {{ESC}:
  Redesigning {WSD} with extractive sense comprehension}.
\newblock In \emph{Proceedings of the 2021 Conference of the North American
  Chapter of the Association for Computational Linguistics: Human Language
  Technologies}, pages 4661--4672, Online. Association for Computational
  Linguistics.

\bibitem[{Blevins and Zettlemoyer(2020)}]{BEM}
Terra Blevins and Luke Zettlemoyer. 2020.
\newblock \href {https://doi.org/10.18653/v1/2020.acl-main.95} {Moving down the
  long tail of word sense disambiguation with gloss informed bi-encoders}.
\newblock In \emph{Proceedings of the 58th Annual Meeting of the Association
  for Computational Linguistics}, pages 1006--1017, Online. Association for
  Computational Linguistics.

\bibitem[{Chawla et~al.(2002)Chawla, Bowyer, Hall, and
  Kegelmeyer}]{oversampling}
N.~V. Chawla, K.~W. Bowyer, L.~O. Hall, and W.~P. Kegelmeyer. 2002.
\newblock \href {https://doi.org/10.1613/jair.953} {{SMOTE}: Synthetic minority
  over-sampling technique}.
\newblock \emph{Journal of Artificial Intelligence Research}, 16:321--357.

\bibitem[{Chou et~al.(2020)Chou, Chang, Pan, Wei, and Juan}]{Remix}
Hsin{-}Ping Chou, Shih{-}Chieh Chang, Jia{-}Yu Pan, Wei Wei, and Da{-}Cheng
  Juan. 2020.
\newblock \href {http://arxiv.org/abs/2007.03943} {Remix: Rebalanced mixup}.
\newblock \emph{CoRR}, abs/2007.03943.

\bibitem[{Galdran et~al.(2021)Galdran, Carneiro, and Ballester}]{BalancedMixup}
Adrian Galdran, Gustavo Carneiro, and Miguel {\'{A}}ngel~Gonz{\'{a}}lez
  Ballester. 2021.
\newblock \href {http://arxiv.org/abs/2109.09850} {Balanced-mixup for highly
  imbalanced medical image classification}.
\newblock \emph{CoRR}, abs/2109.09850.

\bibitem[{He et~al.(2019)He, Xie, Chen, Zhang, Wang, and
  Tian}]{distributionGap}
Zhuoxun He, Lingxi Xie, Xin Chen, Ya~Zhang, Yanfeng Wang, and Qi~Tian. 2019.
\newblock \href {http://arxiv.org/abs/1909.09148} {Data augmentation revisited:
  Rethinking the distribution gap between clean and augmented data}.
\newblock \emph{CoRR}, abs/1909.09148.

\bibitem[{Huang et~al.(2019)Huang, Sun, Qiu, and Huang}]{GlossBERT}
Luyao Huang, Chi Sun, Xipeng Qiu, and Xuanjing Huang. 2019.
\newblock \href {http://arxiv.org/abs/1908.07245} {Glossbert: {BERT} for word
  sense disambiguation with gloss knowledge}.
\newblock \emph{CoRR}, abs/1908.07245.

\bibitem[{Kabra et~al.(2020)Kabra, Chopra, Puri, Badjatiya, Verma, Gupta, and
  Krishnamurthy}]{MixBoost}
Anubha Kabra, Ayush Chopra, Nikaash Puri, Pinkesh Badjatiya, Sukriti Verma,
  Piyush Gupta, and Balaji Krishnamurthy. 2020.
\newblock \href {http://arxiv.org/abs/2009.01571} {Mixboost: Synthetic
  oversampling with boosted mixup for handling extreme imbalance}.
\newblock \emph{CoRR}, abs/2009.01571.

\bibitem[{Kilgarriff(2004)}]{zipfian}
Adam Kilgarriff. 2004.
\newblock How dominant is the commonest sense of a word?
\newblock In \emph{Text, Speech and Dialogue}, pages 103--111, Berlin,
  Heidelberg. Springer Berlin Heidelberg.

\bibitem[{Lee et~al.(2020)Lee, Zaheer, Astrid, and Lee}]{SmoothMix}
Jin-Ha Lee, Muhammad~Zaigham Zaheer, Marcella Astrid, and Seung-Ik Lee. 2020.
\newblock Smoothmix: A simple yet effective data augmentation to train robust
  classifiers.
\newblock In \emph{Proceedings of the IEEE/CVF Conference on Computer Vision
  and Pattern Recognition (CVPR) Workshops}.

\bibitem[{Liu et~al.(2021)Liu, Wang, Shen, Qi, and Tian}]{twostep}
Jiabin Liu, Bo~Wang, Xin Shen, Zhiquan Qi, and Yingjie Tian. 2021.
\newblock \href {http://arxiv.org/abs/2105.10635} {Two-stage training for
  learning from label proportions}.
\newblock \emph{CoRR}, abs/2105.10635.

\bibitem[{Maru et~al.(2022)Maru, Conia, Bevilacqua, and Navigli}]{microF1}
Marco Maru, Simone Conia, Michele Bevilacqua, and Roberto Navigli. 2022.
\newblock \href {https://doi.org/10.18653/v1/2022.acl-long.324} {{N}ibbling at
  the hard core of {W}ord {S}ense {D}isambiguation}.
\newblock In \emph{Proceedings of the 60th Annual Meeting of the Association
  for Computational Linguistics (Volume 1: Long Papers)}, pages 4724--4737,
  Dublin, Ireland. Association for Computational Linguistics.

\bibitem[{Miller et~al.(1993)Miller, Leacock, Tengi, and Bunker}]{semcor}
George~A. Miller, Claudia Leacock, Randee Tengi, and Ross~T. Bunker. 1993.
\newblock \href {https://doi.org/10.3115/1075671.1075742} {A semantic
  concordance}.
\newblock In \emph{Proceedings of the Workshop on Human Language Technology},
  HLT '93, page 303–308, USA. Association for Computational Linguistics.

\bibitem[{Moro and Navigli(2015)}]{SE15}
Andrea Moro and Roberto Navigli. 2015.
\newblock \href {https://doi.org/10.18653/v1/S15-2049} {{S}em{E}val-2015 task
  13: Multilingual all-words sense disambiguation and entity linking}.
\newblock In \emph{Proceedings of the 9th International Workshop on Semantic
  Evaluation ({S}em{E}val 2015)}, pages 288--297, Denver, Colorado. Association
  for Computational Linguistics.

\bibitem[{Navigli et~al.(2013)Navigli, Jurgens, and Vannella}]{SE13}
Roberto Navigli, David Jurgens, and Daniele Vannella. 2013.
\newblock \href {https://aclanthology.org/S13-2040} {{S}em{E}val-2013 task 12:
  Multilingual word sense disambiguation}.
\newblock In \emph{Second Joint Conference on Lexical and Computational
  Semantics (*{SEM}), Volume 2: Proceedings of the Seventh International
  Workshop on Semantic Evaluation ({S}em{E}val 2013)}, pages 222--231, Atlanta,
  Georgia, USA. Association for Computational Linguistics.

\bibitem[{Palmer et~al.(2001)Palmer, Fellbaum, Cotton, Delfs, and Dang}]{SE2}
Martha Palmer, Christiane Fellbaum, Scott Cotton, Lauren Delfs, and Hoa~Trang
  Dang. 2001.
\newblock \href {https://aclanthology.org/S01-1005} {{E}nglish tasks: All-words
  and verb lexical sample}.
\newblock In \emph{Proceedings of {SENSEVAL}-2 Second International Workshop on
  Evaluating Word Sense Disambiguation Systems}, pages 21--24, Toulouse,
  France. Association for Computational Linguistics.

\bibitem[{Pradhan et~al.(2007)Pradhan, Loper, Dligach, and Palmer}]{se07}
Sameer Pradhan, Edward Loper, Dmitriy Dligach, and Martha Palmer. 2007.
\newblock \href {https://aclanthology.org/S07-1016} {{S}em{E}val-2007 task-17:
  {E}nglish lexical sample, {SRL} and all words}.
\newblock In \emph{Proceedings of the Fourth International Workshop on Semantic
  Evaluations ({S}em{E}val-2007)}, pages 87--92, Prague, Czech Republic.
  Association for Computational Linguistics.

\bibitem[{Raffel et~al.(2019)Raffel, Shazeer, Roberts, Lee, Narang, Matena,
  Zhou, Li, and Liu}]{T5}
Colin Raffel, Noam Shazeer, Adam Roberts, Katherine Lee, Sharan Narang, Michael
  Matena, Yanqi Zhou, Wei Li, and Peter~J. Liu. 2019.
\newblock \href {http://arxiv.org/abs/1910.10683} {Exploring the limits of
  transfer learning with a unified text-to-text transformer}.
\newblock \emph{CoRR}, abs/1910.10683.

\bibitem[{Snyder and Palmer(2004)}]{SE3}
Benjamin Snyder and Martha Palmer. 2004.
\newblock \href {https://aclanthology.org/W04-0811} {The {E}nglish all-words
  task}.
\newblock In \emph{Proceedings of {SENSEVAL}-3, the Third International
  Workshop on the Evaluation of Systems for the Semantic Analysis of Text},
  pages 41--43, Barcelona, Spain. Association for Computational Linguistics.

\bibitem[{Taghipour and Ng(2015)}]{OMSTI}
Kaveh Taghipour and Hwee~Tou Ng. 2015.
\newblock \href {https://doi.org/10.18653/v1/K15-1037} {One million
  sense-tagged instances for word sense disambiguation and induction}.
\newblock In \emph{Proceedings of the Nineteenth Conference on Computational
  Natural Language Learning}, pages 338--344, Beijing, China. Association for
  Computational Linguistics.

\bibitem[{Van Der~Maaten(2014)}]{tsne}
Laurens Van Der~Maaten. 2014.
\newblock Accelerating t-sne using tree-based algorithms.
\newblock \emph{J. Mach. Learn. Res.}, 15(1):3221–3245.

\bibitem[{Wu et~al.(2020)Wu, Zhang, Valiant, and
  R{\'{e}}}]{LinearTransformationDataAug}
Sen Wu, Hongyang~R. Zhang, Gregory Valiant, and Christopher R{\'{e}}. 2020.
\newblock \href {http://arxiv.org/abs/2005.00695} {On the generalization
  effects of linear transformations in data augmentation}.
\newblock \emph{CoRR}, abs/2005.00695.

\bibitem[{Yoon et~al.(2021)Yoon, Kim, and Park}]{SSMix}
Soyoung Yoon, Gyuwan Kim, and Kyumin Park. 2021.
\newblock \href {http://arxiv.org/abs/2106.08062} {Ssmix: Saliency-based span
  mixup for text classification}.
\newblock \emph{CoRR}, abs/2106.08062.

\bibitem[{Zhang et~al.(2017)Zhang, Ciss{\'{e}}, Dauphin, and
  Lopez{-}Paz}]{mixup}
Hongyi Zhang, Moustapha Ciss{\'{e}}, Yann~N. Dauphin, and David Lopez{-}Paz.
  2017.
\newblock \href {http://arxiv.org/abs/1710.09412} {mixup: Beyond empirical risk
  minimization}.
\newblock \emph{CoRR}, abs/1710.09412.

\end{thebibliography}
\bibliographystyle{acl_natbib}

\newpage
\appendix
\section{Appendix}
\label{sec:appendix}
\subsection{t-SNE Plots}
\label{tsneplots}
\begin{figure}[h]
 \centering
 \vspace*{5pt}%
 \hspace*{\fill}%
  \begin{subfigure}{\linewidth}     
    \centering
    \includegraphics[width=\textwidth]{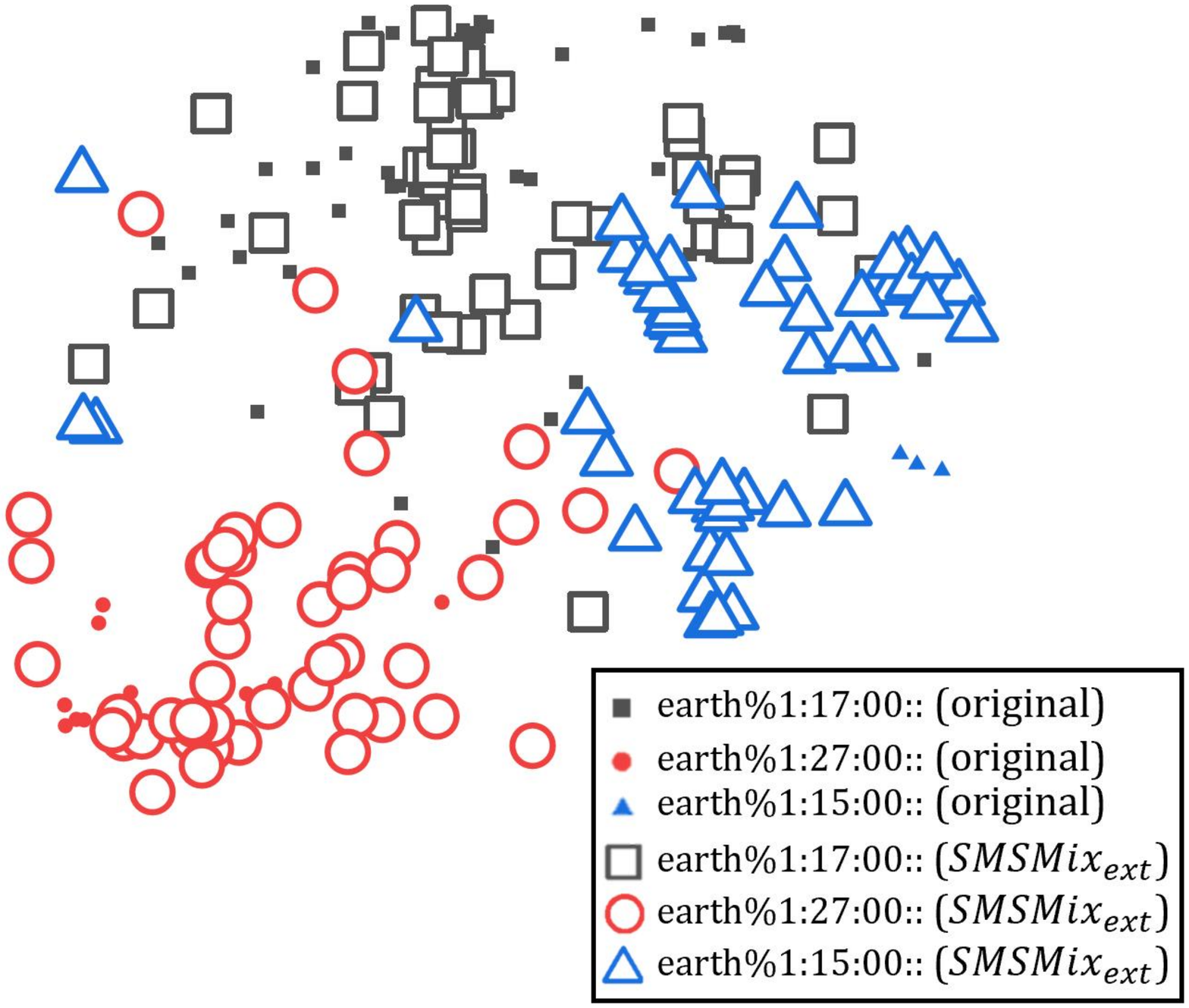}%
    \captionsetup{skip=12pt}%
    \caption{earth (noun)}
    \label{fig:tsne-earth}
  \end{subfigure}
  \hspace*{\fill}

  \vspace*{8pt}%

  \hspace*{\fill}%
   \begin{subfigure}{\linewidth}        
    \centering
    \includegraphics[width=\textwidth]{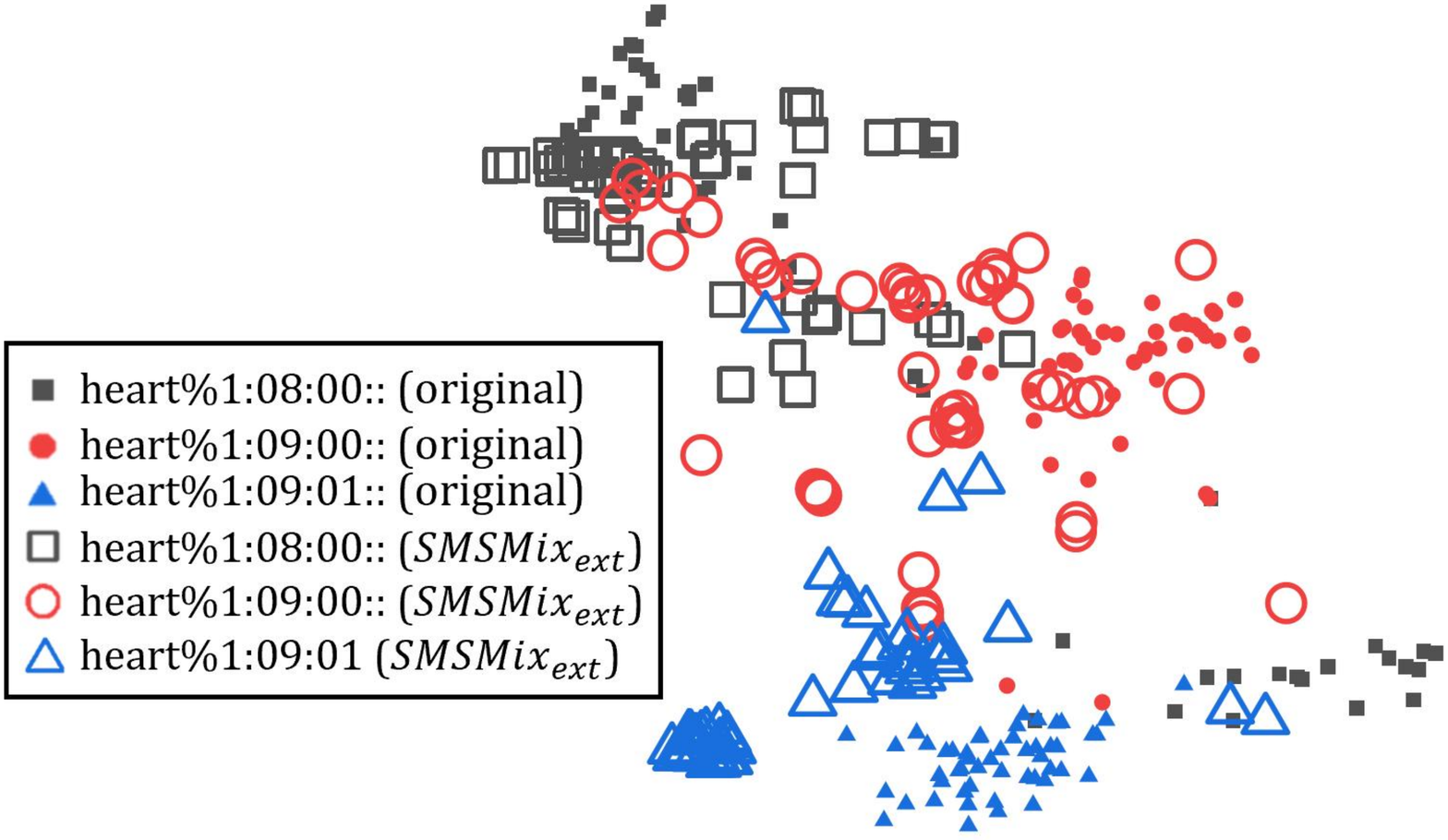}%
    \captionsetup{skip=12pt}%
    \caption{heart (noun)}
    \label{fig:tsne-heart}
  \end{subfigure}
  \hspace*{\fill}

  \vspace*{8pt}%

  \hspace*{\fill}%
  \begin{subfigure}{\linewidth}     
    \centering
    \includegraphics[width=\textwidth]{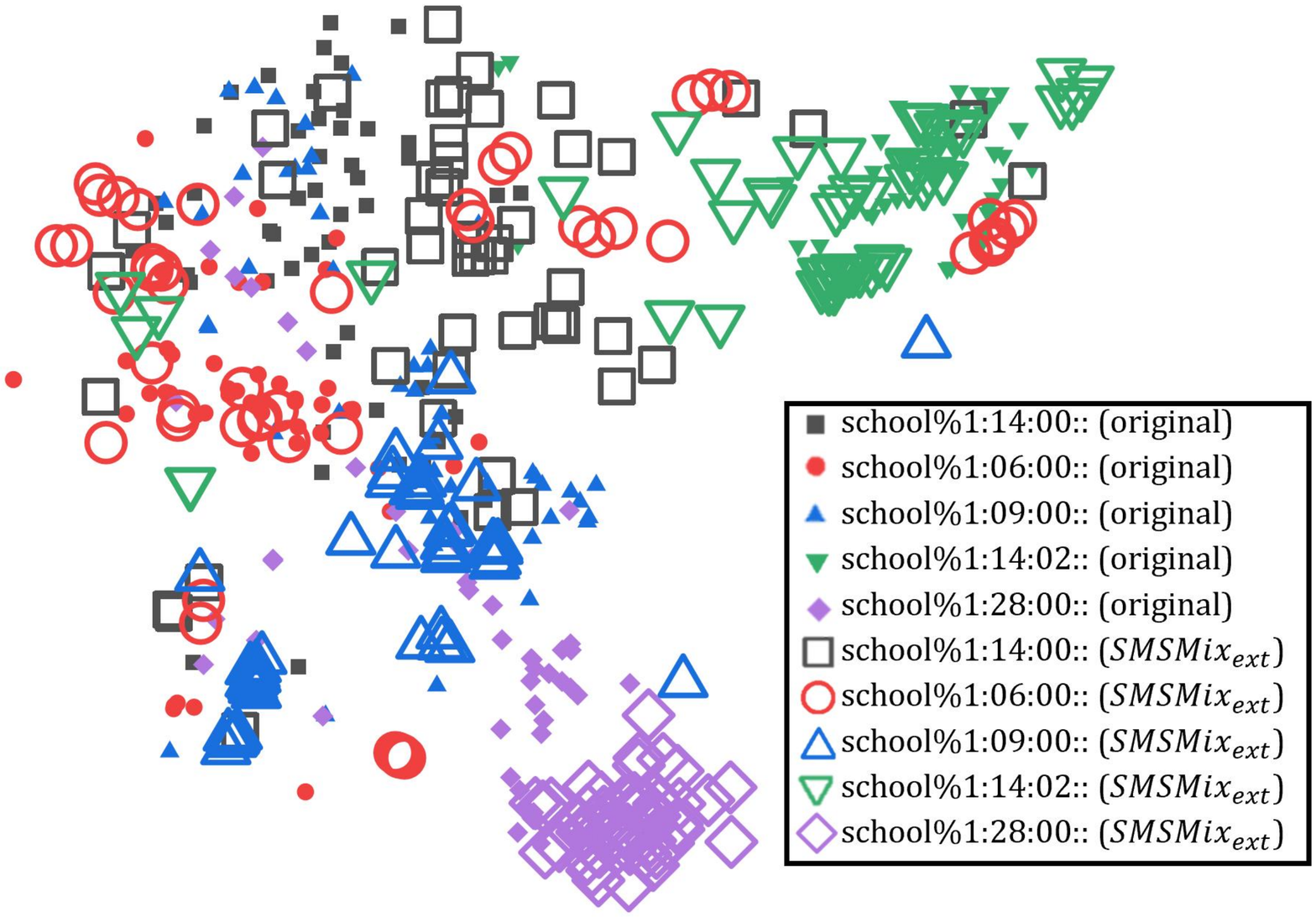}%
    \captionsetup{skip=12pt}%
    \caption{school (noun)}
    \label{fig:demo3}
  \end{subfigure}
  \hspace*{\fill}%
  \caption{Latent space visualization showing that SMSMix maintains the sense lablel of the target word.}
  \label{fig:tsnes}
\end{figure}

We present additional t-SNE plots in Figure~\ref{fig:tsnes} to show that the augmentation data generated by our method retain the label of the original data. The method of obtaining the t-SNE plots is the same as in section \ref{sec:tsne}.

\subsection{Augmentation Examples}
\label{aug_example}
We show examples of augmented sentences for several senses in Tables \ref{tab:good} and \ref{tab:fail}. Table \ref{tab:good} shows some examples of well-augmented results using our proposed SMSMix. The chosen span with a target word, determined by the saliency score from the original sentence, smoothly blends in with a randomly sampled Wikipedia sentence. On the other hand, the examples shown in Table \ref{tab:fail} are cherry-picked failure examples. There are a few failure cases where there are incorrect grammar uses or the sentence does not make sense due to the high discrepancy of context between the two sentences.

\subsection{Standard Deviation of Results}
\label{std}
In Table \ref{tab:overallstd} and Table \ref{tab:freqstd}, we report the standard deviation of the results obtained in Table \ref{tab:overall} and Table \ref{tab:freq}, respectively. The values were obtained by running the same experiments with five different random seeds.

\subsection{Location for Span Injection}
\label{spaninjection}
\begin{figure*}[hbt!]
	\centering
	\includegraphics[width=\linewidth]{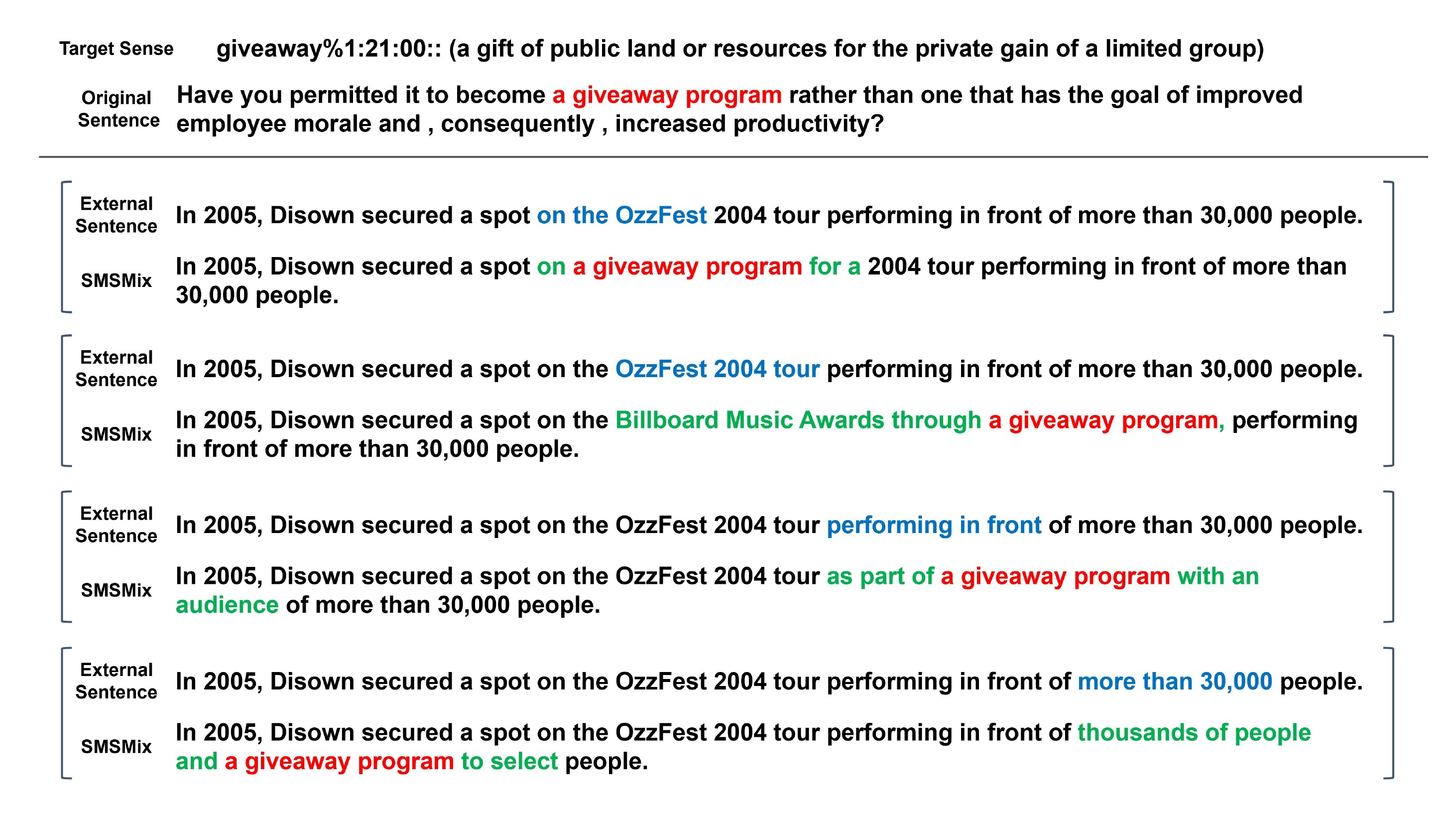}
	\caption{Examples of SMSMix using various different injection locations. \textcolor{red}{Red} represents the sense-maintained span in the original sentence, \textcolor{blue}{blue} represents a random span in an external sentence that is going to be replaced, and \textcolor{green}{green} represents the span prediction made by T5.}
	\label{fig:appendix2}
\end{figure*}

We randomly choose where to inject the sense-maintained span in a sentence, as we use T5's span prediction capability to complete the sentence. To demonstrate the effectiveness of T5's span prediction for SMSMix, we use the example in Figure \ref{fig:2} to create various augmented sentences by injecting the sense-maintained span into different random locations (Figure \ref{fig:appendix2}). 

\begin{table*}[hbt!]

	\centering
	\begin{tabular}[t]{l c c|| c c|}
		\Xhline{3\arrayrulewidth}
		\multicolumn{2}{c}{Examples}     \\ 
		\Xhline{2\arrayrulewidth}
		\bf{target sense}  & \makecell[l]{output\%1:10:02:: (signal that comes out of an electronic system)} \\
        \bf{original}  & \makecell[tl]{\textit{\textcolor{red}{\textbf{Outputs} of the two systems are measured}} by a pulse timing circuit and a resistance \\bridge, followed by a simple analogue computer which feeds a multichannel recorder.}\\
        \bf{Wikipedia} & \makecell[tl]{The book tells the story of \textcolor{blue}{Hendrix and his life through reproductions of} rare material\\ such as letters, drawings, postcards and posters.}\\
        \Xhline{0.5\arrayrulewidth}
        \bf{SMSMix}  & \makecell[tl]{The book tells the story of \textcolor{green}{how the inputs and} \textit{\textcolor{red}{\textbf{outputs} of the two systems are}}\\ \textit{\textcolor{red}{measured}} \textcolor{green}{through} rare material such as letters, drawings, postcards and posters.}\\
        
        \hline
        \hline
        
        \bf{target sense}  & \makecell[l]{work\%1:04:01:: (the occupation for which you are paid) } \\
        \bf{original}  & \makecell[tl]{	In a few places cooperative programs between schools and employers in \textit{\textcolor{red}{clerical}}\\ \textit{\textcolor{red}{\textbf{work}}} \textit{\textcolor{red}{have}} shown the same possibilities for allowing the student, while still in \\ school, to develop skills which are immediately marketable upon graduation.}\\
        \bf{Wikipedia} & \makecell[tl]{Throughout his career, \textcolor{blue}{he was the} recipient of more than 30 awards and honors \\related to engineering, manufacturing, and the development of heavy equipment.}\\
        \Xhline{0.5\arrayrulewidth}
        \bf{SMSMix}  & \makecell[tl]{Throughout his career, \textcolor{green}{his engineering skills and} \textit{\textcolor{red}{clerical \textbf{work} have}} \textcolor{green}{been the} \\ recipient of more than 30 awards and honors related to engineering, manufacturing \\,and the development of heavy equipment.}\\
        \hline
        \hline
        \bf{target sense}  & \makecell[l]{work\%2:38:00:: (proceed along a path)  } \\
        \bf{original}  & \makecell[tl]{	Several photographs and charts of galaxies help the non-scientist keep up \\with the discussion, and the smooth language indicates the contributors were\\ determined to avoid the \textit{\textcolor{red}{jargon that seems to \textbf{work} its way into}} almost every field.}\\
        \bf{Wikipedia} & \makecell[tl]{Barbour gave his time free for the next 25 summers \textcolor{blue}{to manage field parties }\\ \textcolor{blue}{throughout the state}, surveying the geological and paleontological resources \\ of the State of Nebraska.}\\
        \Xhline{2 \arrayrulewidth}
        \bf{SMSMix}  & \makecell[tl]{Barbour gave his time free for the next 25 summers \textcolor{green}{to deal with all the} \textit{\textcolor{red}{jargon that}}\\ \textit{\textcolor{red}{seems to \textbf{work} its way into}} \textcolor{green}{the process of} surveying the geological and \\ paleontological resources of the State of Nebraska.}\\
        \hline
        \hline
        \bf{target sense}  & \makecell[l]{condition\%1:10:01:: \\(an assumption on which rests the validity or effect of something else) } \\
        \bf{original}  & \makecell[tl]{	This is what we mean when we say this demand must be accepted \textit{\textcolor{red}{without \textbf{condition.}}}}\\
        \bf{Wikipedia} & \makecell[tl]{The Society \textcolor{blue}{was wound up} in the year 2001 when no ordinary members wanted to be\\ nominated as new committee members.}\\
        \Xhline{2 \arrayrulewidth}
        \bf{SMSMix}  & \makecell[tl]{The Society \textcolor{green}{is open to all,} \textit{\textcolor{red}{without \textbf{condition}}} \textcolor{green}{, except} in the year 2001 when no \\ ordinary members wanted to be nominated as new committee members. }\\
        \hline
        \hline
        \bf{target sense}  & \makecell[l]{lighting\%1:06:00:: \\(apparatus for supplying artificial light effects for the stage or a film) } \\
        \bf{original}  & \makecell[tl]{		When improvements are recommended in \textit{\textcolor{red}{working conditions - such as \textbf{lighting}}}\\ , \textit{\textcolor{red}{rest rooms}} , eating facilities , air-conditioning - do you try to set a measure of\\ their effectiveness on productivity?}\\
        \bf{Wikipedia} & \makecell[tl]{	However, when de Gaulle first introduced \textcolor{blue}{the Fouchet Plan in 1961, it faced} \\  \textcolor{blue}{opposition from many of the} member states.}\\
        \Xhline{2 \arrayrulewidth}
        \bf{SMSMix}  & \makecell[tl]{However, when de Gaulle first introduced \textcolor{green}{the idea, improvements are recommended}\\ \textcolor{green}{in} \textit{\textcolor{red}{working conditions - such as \textbf{lighting}, rest rooms}} \textcolor{green}{and transport between} \\ member states.}\\

      \Xhline{2\arrayrulewidth}

	\end{tabular}
	\caption{Examples of well-augmented sentences, generated by SMSMix. \textcolor{red}{Red} represents the sense-maintained span in the original sentence, \textcolor{blue}{blue} represents a random span in an external sentence that is going to be replaced, and \textcolor{green}{green} represents the span prediction made by T5. }
	\label{tab:good}
\end{table*}

\begin{table*}[hbt!]

	\centering
	\begin{tabular}[t]{l c c|| c c|}
		\Xhline{3\arrayrulewidth}
		\multicolumn{2}{c}{Examples}     \\ 
		\Xhline{2\arrayrulewidth}
		\bf{target sense}  & \makecell[l]{make\%2:40:02:: (achieve a point or goal)} \\
        \bf{original}  & \makecell[tl]{It is interesting to note that medium \textit{\textcolor{red}{compulsives in the unstructured schools \textbf{made}}} \\ \textit{\textcolor{red}{lowest achievement scores}} ( although not significantly lower ) .}\\
        \bf{Wikipedia} & \makecell[tl]{Josh tries to distance himself as much as possible from her, for fear of \\\textcolor{blue}{what might happen if she finds out what he is.}}\\
        \Xhline{0.5\arrayrulewidth}
        \bf{SMSMix}  & \makecell[tl]{Josh tries to distance himself as much as possible from her, for fear of \textcolor{green}{losing her,}\\ \textcolor{green}{but the} \textit{\textcolor{red}{compulsives in the unstructured schools \textbf{made} the lowest achievement scores}}\\ \textcolor{green}{in the}.}\\
        \hline
        \hline
        \bf{target sense}  & \makecell[l]{employment\%1:04:01:: (the act of using) } \\
        \bf{original}  & \makecell[tl]{Another case may be given in illustration of a successful use \textit{\textcolor{red}{of analysis, and }}\\\textit{\textcolor{red}{also of the \textbf{employment} of a procedure}} for intensive analysis.}\\
        \bf{Wikipedia} & \makecell[tl]{Ahh!\textcolor{blue}{, which was released on May 21, 1988 and} would ultimately go on to sell \\8 million copies worldwide.}\\
        \Xhline{2 \arrayrulewidth}
        \bf{SMSMix}  & \makecell[tl]{Ahh! \textcolor{red}{\textit{of analysis, and also of the \textbf{employment} of a procedure}} \textcolor{green}{that} would ultimately \\go on to sell 8 million copies worldwide.}\\
        \hline
        \hline
        \bf{target sense}  & \makecell[l]{replace\%2:41:00:: (take the place or move into the position of)} \\
        \bf{original}  & \makecell[tl]{	This and raw \textcolor{red}{\textit{sugar \textbf{replace} ordinary}} refined sugar on the tables and \\very little sugar is used in cooking.}\\
        \bf{Wikipedia} & \makecell[tl]{These are public housing units and estates aimed at Singaporeans who \textcolor{blue}{do not want} \\a HDB flat but might  find private property too expensive.}\\
        \Xhline{0.5\arrayrulewidth}
        \bf{SMSMix}  & \makecell[tl]{These are public housing units and estates aimed at Singaporeans who \textcolor{green}{want to} \textcolor{red}{\textit{sugar }}\\\textcolor{red}{\textit{\textbf{replace} ordinary}} \textcolor{green}{sugar} a HDB flat but might find private property too expensive.}\\
        \hline
        \hline
        \bf{target sense}  & \makecell[l]{people\%1:14:03:: (the common people generally)} \\
        \bf{original}  & \makecell[tl]{Linguists have not always been more enlightened than ``\textcolor{red}{\textit{practical \textbf{people}'' and }}\\\textcolor{red}{\textit{sometimes have insisted on incredibly trivial}} points while neglecting things\\ of much greater significance.}\\
        \bf{Wikipedia} & \makecell[tl]{According to the law of marginal utility, the value of each good in a stock of \\identical goods is \textcolor{blue}{utility of the last and most easily dispensable unit}.}\\
        \Xhline{2 \arrayrulewidth}
        \bf{SMSMix}  & \makecell[tl]{According to the law of marginal utility , the value of each good in a stock\\ of identical goods is \textcolor{green}{``the most} \textcolor{red}{\textit{practical \textbf{people}'' and sometimes have insisted on}}\\ \textcolor{red}{\textit{incredibly trivial}} \textcolor{green}{details}.}\\
        
        \hline
        \hline
        \bf{target sense}  & \makecell[l]{shift\%2:38:02:: (move around)} \\
        \bf{original}  & \makecell[tl]{Important as these differences are, they should not obscure the basic fact that \textcolor{red}{\textit{by}} \\\textcolor{red}{\textit{\textbf{shifting} the hypothalamic balance}} sufficiently to the parasympathetic side, we\\ produce depressions, whereas a shift in the opposite direction causes excitatory \\ effects and, eventually, maniclike changes.}\\
        \bf{Wikipedia} & \makecell[tl]{The district is the mining and forestry centre of Suriname, with many \textcolor{blue}{large} \\\textcolor{blue}{bauxite mining operations operating}.}\\
        \Xhline{2 \arrayrulewidth}
        \bf{SMSMix}  & \makecell[tl]{The district is the mining and forestry centre of Suriname, with many \textcolor{green}{mines} \\\textcolor{red}{\textit{by \textbf{shifting} the hypothalamic balance}} \textcolor{green}{of}.}\\
      
       \Xhline{2\arrayrulewidth}

	\end{tabular}
	\caption{Failure examples of augmentation by SMSMix. \textcolor{red}{Red} represents the sense-maintained span in the original sentence, \textcolor{blue}{blue} represents a random span in an external sentence that is going to be replaced, and \textcolor{green}{green} represents the span prediction made by T5. }
	\label{tab:fail}
\end{table*}

\begin{table*}[hbt!]
	\centering
	\begin{tabular}{l||c c| c c c c c c c c c c}
		\Xhline{3\arrayrulewidth}
		& \multicolumn{2}{c|}{Dev Set} & \multicolumn{10}{c}{Test Sets} \\
		\cline{2-13}
		Methods                              & \multicolumn{2}{c|}{SE07}     & \multicolumn{2}{c}{SE2}     &\multicolumn{2}{c}{SE3}     & \multicolumn{2}{c}{SE13}     & \multicolumn{2}{c}{SE15}   & \multicolumn{2}{c}{ALL}     \\  
		                                    &\begin{small}m-F1\end{small} &\begin{small}M-F1\end{small}&\begin{small}m-F1\end{small} &\begin{small}M-F1\end{small}&\begin{small}m-F1\end{small} &\begin{small}M-F1\end{small}&\begin{small}m-F1\end{small} &\begin{small}M-F1\end{small}&\begin{small}m-F1\end{small} &\begin{small}M-F1\end{small}&\begin{small}m-F1\end{small} &\begin{small}M-F1\end{small}\\ 
		\Xhline{2\arrayrulewidth}
		\begin{small} +Oversample\end{small}   & 0.38 & 0.41 & 0.10 & 0.10 & 0.11 & 0.18 & 0.10 & 0.20 & 0.06 & 0.10 & 0.49 & 0.10 \\ 
		\begin{small} +SMSMix$_{int}$\end{small} & 0.15 & 0.18 & 0.09 & 0.11 & 0.18 & 0.20 & 0.23 & 0.30 & 0.10 & 0.06 & 0.05 & 0.05\\
		\begin{small} +SMSMix$_{ext}$\end{small} & 0.19 & 0.22 & 0.04 & 0.09 & 0.08 & 0.06 & 0.09 & 0.15 & 0.08 & 0.09 & 0.05 & 0.05
		\\\hline

		\Xhline{3\arrayrulewidth} 

	\end{tabular}
	\caption{Standard deviation values for the experimental results in Table \ref{tab:overall}. The values were obtained by running the same experiments with five different random seeds.}
	\label{tab:overallstd}
\end{table*}

\begin{table*}[hbt!]
	\centering

	\begin{tabular}{l|| c c c c c c c c c c}
		\Xhline{3\arrayrulewidth}
		  &\multicolumn{2}{c}{MFS}     & \multicolumn{2}{c}{LFS}     & \multicolumn{2}{c}{0-lex}   & \multicolumn{2}{c}{0-lex-def}  &\multicolumn{2}{c}{0-def}   \\  
		                                    &\begin{small}m-F1\end{small} &\begin{small}M-F1\end{small}&\begin{small}m-F1\end{small} &\begin{small}M-F1\end{small}&\begin{small}m-F1\end{small} &\begin{small}M-F1\end{small}&\begin{small}m-F1\end{small} &\begin{small}M-F1\end{small}&\begin{small}m-F1\end{small} &\begin{small}M-F1\end{small}\\ 
		\Xhline{2\arrayrulewidth}
		+ Oversample & 0.11 & 0.17  & 0.14 & 0.15 & 0.12 & 0.15 & 0.15 & 0.13 & 0.18 &0.25\\
		+ SMSMix$_{int}$ & 0.07 & 0.06 & 0.19 & 0.10 & 0.04 & 0.10 & 0.10 & 0.13 & 0.10 & 0.13\\
		+ SMSMix$_{ext}$  & 0.08 & 0.04 & 0.12 & 0.12 & 0.04 & 0.08 & 0.10 & 0.07 & 0.08 & 0.07\\

		\Xhline{3\arrayrulewidth}
	\end{tabular}
	\caption{Standard deviation values for the experimental results in Table \ref{tab:freq}. The values were obtained by running the same experiments with five different random seeds.}
	\label{tab:freqstd}
\end{table*}

\end{document}